\pdfoutput=1
\documentclass[11pt]{article}

\usepackage[preprint]{acl}
\usepackage{float}
\usepackage{placeins}

\usepackage{times}
\usepackage{tabularx}
\usepackage{latexsym}
\usepackage{float}
\usepackage{todonotes}
\usepackage{enumerate}
\usepackage{enumitem}
\usepackage{amsmath}
\usepackage[T1]{fontenc}
\usepackage[utf8]{inputenc}

\usepackage{microtype}

\usepackage{inconsolata}

\usepackage{graphicx}
\usepackage{booktabs}
\usepackage{hyperref}

\title{FRaN-X: FRaming and Narratives-eXplorer}

\author{
Artur Muratov$^{1,2^*}$, 
Hana Fatima Shaikh$^{1,3^*}$, 
Vanshikaa Jani$^{1,4^*}$, 
Tarek Mahmoud$^1$, \\
\textbf{Zhuohan Xie}$^1$,
\textbf{Daniil Orel}$^1$,
\textbf{Aaryamonvikram Singh}$^1$,
\textbf{Yuxia Wang}$^1$, 
\textbf{Aadi Joshi}$^1$, \\
\textbf{Hasan Iqbal}$^1$,
\textbf{Ming Shan Hee}$^1$,
\textbf{Dhruv Sahnan}$^1$,
\textbf{Nikolaos Nikolaidis}$^5$, \\
\textbf{Purificação Silvano}$^6$, 
\textbf{Dimitar Dimitrov}$^7$, 
\textbf{Roman Yangarber}$^8$
\textbf{Ricardo Campos}$^{9}$, \\
\textbf{Alípio Jorge}$^{10}$, 
\textbf{Nuno Guimarães}$^{10}$,
\textbf{Elisa Sartori}$^{11}$,  
\textbf{Nicolas Stefanovitch}$^{12}$, \\
\textbf{Giovanni Da San Martino}$^{11}$,
\textbf{Jakub Piskorski}$^{13}$, 
\textbf{Preslav Nakov}$^1$\\
$^1$MBZUAI, UAE \quad
$^2$Nazarbayev University, Kazakhstan \quad
$^3$University of Maryland, USA \\
$^4$University of Arizona, USA 
$^5$Athens University of Economics and Business, Greece \\
$^6$University of Porto, Portugal \quad
$^7$Sofia University "St. Kliment Ohridski", Bulgaria \\
$^8$University of Helsinki, Finland \quad
$^{9}$Beira Interior and INESC TEC, Portugal \\
$^{10}$Porto and INESC TEC, Portugal \quad
$^{11}$University of Padova, Italy \\
$^{12}$European Commission Joint Research Center, Italy \\
$^{13}$Polish Academy of Sciences, Poland\\
}

\begin{document}

\maketitle

\def\thefootnote{*}\footnotetext{Equal contribution.}\def\thefootnote{\arabic{footnote}}

\begin{abstract}
We present \mbox{FRaN-X}\footnote{This work was done during a summer internship at the NLP department, MBZUAI.}, a Framing and Narratives Explorer that detects entity mentions and classifies their narrative roles directly from raw text. \mbox{FRaN-X} is a two-stage system that combines sequence labeling with fine-grained role classification to reveal how entities are portrayed as protagonists, antagonists, or innocents, with a taxonomy of 22 fine-grained roles nested under these three main categories. The system supports five languages (Bulgarian, English, Hindi, Russian, and Portuguese) and two domains (the Russia--Ukraine War and Climate Change). It provides an interactive web interface for media analysts to explore and compare framing across different sources, tackling the challenge of detecting and labeling how entities are framed. Our system allows end-users to focus on a single article as well as analyze up to four articles simultaneously. We provide aggregate level analysis including an intuitive graph visualization that highlights the narrative a group of articles are pushing. Our system includes a search feature for users to look up entities of interest, along with a timeline view that allows analysts to track an entity’s role transitions across different contexts within the article. The FRaN-X system and the trained models are licensed under an MIT License. \mbox{FRaN-X} is publicly accessible at \url{https://fran-x.streamlit.app/} and a video demonstration is available at \url{https://youtu.be/VZVi-1B6yYk}. %\todo{Update the video link before submission link. You can do a draft version before overriding with the final version} 

% \textcolor{red}{Video Link Not Working}
\end{abstract}

\section{Introduction}

News narratives are shaped not just by facts but also by how entities are framed within stories. Coverage of the same event often varies across outlets. This reflected by differing perspectives, biases, and persuasion strategies. This makes it challenging for readers to distinguish factual reporting from opinion or bias. As a result, there is a growing need for tools that can support systematic analysis of news articles to allow readers to understand any potential biases and to equip researchers and engineers with the necessary tools to research and develop systems that counter such bias.

Several tools have been developed to analyze news from different perspectives. For instance, Media Cloud~\citep{reoberts2021mediacloud} enables users to track media attention over time. Voyant Tools~\citep{sinclair_rockwell_voyant_2016} provides general-purpose text analytics for exploratory news analysis. NewsLens~\citep{laban-hearst-2017-newslens} focuses on constructing story threads spanning years and sources. More relevant to our work, Tanbih~\citep{zhang-etal-2019-tanbih} profiles thousands of news outlets for bias and factuality, and PRTA~\citep{da-san-martino-etal-2020-prta} allows exploration of propaganda techniques within news articles. More recently, FRAPPE~\citep{sajwani-etal-2024-frappe} provides an online platform for multilingual, and large-scale analysis of framing, persuasion, and propaganda strategies at the \textit{article level}.

However, existing tools primarily focus on \textit{article-level} framing or \textit{outlet-level} profiling, and do not provide insights into the narrative roles assigned to individual entities within stories. To this end, we present FRaN-X (\textbf{FRa}ming and \textbf{N}arratives-e\textbf{X}plorer), an interactive web-based system that performs narrative role framing of central entities. FRaN-X builds on the taxonomy of \citet{mahmoud2025entityframingroleportrayal}, which defines three main roles (Protagonist, Antagonist, Innocent) and 22 fine-grained sub-roles. It enables both on-the-fly analysis of single articles and comparative exploration of framing patterns across multiple documents, providing end-users with unique insights into narrative roles of entities of interest.

\section{System Architecture}

\begin{figure}[t]
  \centering
  \includegraphics[width=1\linewidth]{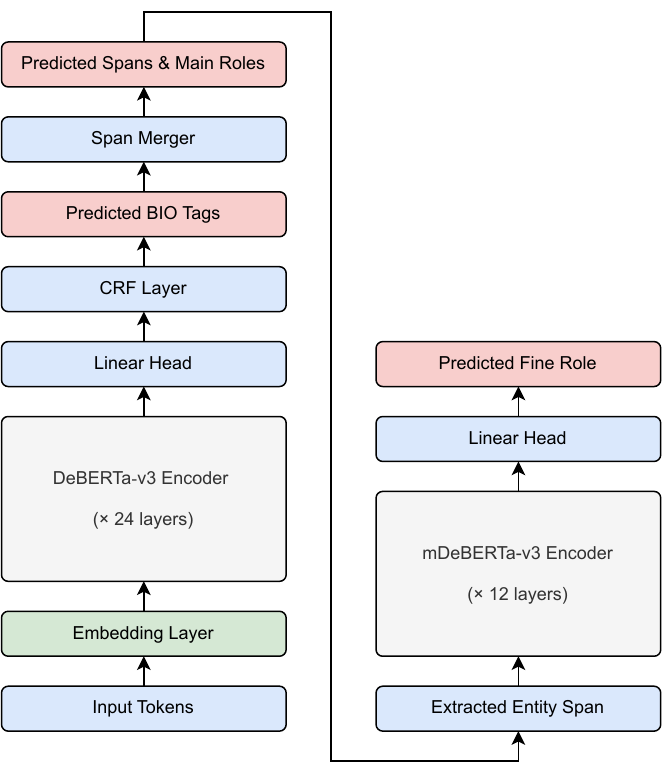}
   \caption{\textbf{The FRaN-X model architecture.} The system processes raw text in a two-stage pipeline: it first performs span detection and main role assignment using sequence labeling, then applies fine-grained role classification to each extracted entity.}
  \label{fig:architecture}
\end{figure}

Figure~\ref{fig:architecture} shows the complete FRaN-X architecture. The system processes raw text input through a two-stage architecture. First, input tokens are embedded and passed through a 24-layer DeBERTa-v3 encoder~\cite{he2021debertav3}, followed by a linear projection layer and a first-order CRF layer~\cite{crf}. This sequence labeling module predicts token-level BIO tags to jointly detect entity spans and assign their main narrative roles. The predicted tags are then post-processed by a custom span merger to reconstruct valid entity mentions with consistent main roles. Each extracted entity span is subsequently fed into a separate 12-layer mDeBERTa-v3  encoder~\cite{he2021debertav3} with a linear classification head that assigns one or more of 22 fine-grained narrative roles that are shown in Appendix~\ref{appendix:taxonomy}. This modular pipeline enables end-to-end entity identification, role framing, and detailed narrative analysis from raw multilingual news text.

\section{Dataset}

We use the dataset provided for Subtask 1 (Entity Framing) in SemEval-2025 Task 10~\citep{piskorski2025semeval}. The dataset covers two domains (the Ukraine--Russia Conflict and Climate Change), and includes articles in five languages (Bulgarian, English, Hindi, Portuguese, and Russian). Each news article contains named entity mentions annotated with spans and labeled with one or more narrative roles, based on a two-level taxonomy of 22 fine-grained roles grouped into three main categories: Protagonist, Antagonist, and Innocent.

While the original task assumes pre-extracted entity mentions, we reformulate it as sequence labeling on raw text, aligning the task more closely with real-world applications. This reformulation increases the task complexity by requiring simultaneous entity detection and role assignment. To reformulate the task into sequence labeling, we converted the training annotated spans into token-level BIO format, so that the model can learn to detect entity mentions and assign them coarse role labels in a single step. More detail about this is reformulation including the various dataset processing methods is given in Section~\ref{sec:seq_lab_data}.

\section{Experiments}

% In our reformulated setting, the system performs full end-to-end sequence labeling. It first detects entity mentions in raw text and predicts their main coarse roles, then classifies each detected span into its fine-grained narrative role.

\subsection{Sequence Labeling}
\label{sec:seq_lab}

To perform entity identification and coarse-role assignment, we cast the task as a single token-level BIO-tagging problem and fine-tune a DeBERTa-v3 with a linear projection and a first-order conditional random field (CRF) layer.

\subsubsection{Data Augmentation}\label{sec:seq_lab_data}
We train on the original gold annotations and explore two data augmentation
methods detailed in Appendix~\ref{app:data_aug}:
\textit{(i) Propagated}, which copies gold labels to co-referent
mentions via surface matching and alias clustering, and
\textit{(ii) Unknown}, which runs a spaCy \cite{honnibal2020spacy} NER pipeline on top of the
Propagated set to label all non-annotated entities as \emph{Unknown}.  
As shown in Appendix~\ref{app:data_aug}, training on the \textit{Propagation}-augmented dataset
yields an absolute 14.3~percentage point gain in exact-match accuracy, while training with the \textit{Unknown}-augmented dataset slightly reduces accuracy. Unless noted otherwise, all results use the best-performing \textit{Propagation}-augmented dataset.

% \subsubsection{Data preparation}\label{sec:seq_lab_data}

% To prepare suitable training data for this sequence labeling setup, we created and evaluated three dataset variants. The first variant (Gold) uses only the original gold annotations without any modifications. The second variant (Propagated) extends the gold labels by automatically propagating them backward and forward within each document through surface-form matching and alias clustering, ensuring that coreferent mentions and name variants receive consistent labels. This propagation follows the original annotation guideline~\cite{mahmoud2025entityframingroleportrayal}, which required annotators to label only the first clear mention of an entity and made repeated mentions optional. The third variant (Unknown) further augments the dataset by automatically extracting additional non-annotated named entities using a spaCy NER pipeline~\cite{honnibal2020spacy} and labeling these with a separate ``Unknown'' class. This variant was designed with the assumption that an explicit Unknown class might help the model distinguish central, narrative-defining entities from peripheral or background mentions, improving framing predictions. However, in our experiments the Unknown variant reduced exact match accuracy and did not yield a clear benefit, as shown in Table~\ref{tab:bert_variant_results}, and was therefore not included in the final system.

\subsubsection{Model Development}

Our model development efforts began with an English-only baseline using BERT-base~\cite{devlin2018bert}, which we used to test the Gold, Propagated, and Unknown dataset variants. We present the results for BERT-base on these variants in Table~\ref{tab:bert_variant_results}. After validating the value of mention propagation and observing the limited value of the \emph{Unknown} class, we trained stronger sequence encoders based on the DeBERTa family. Specifically, we experimented with three configurations: a multilingual DeBERTa-v3-Large, a multilingual mDeBERTa-v3-Base, and a monolingual English-only DeBERTa-v3-Large. These configurations were evaluated across all five target languages, and their comparative performance is summarized in Table~\ref{tab:deberta_comparison}.

While the monolingual DeBERTa-v3-Large achieves the highest score on English (79.1\%), its cross-lingual performance degrades sharply on other languages (e.g., only 7.9\% on Hindi). In contrast, the multilingual DeBERTa-v3-Large offers balanced performance across all languages. Therefore, we selected the multilingual DeBERTa-v3-Large to be our sequence-labeling model.\footnote{The sequence labeling model is publicly available on Hugging Face at \url{https://huggingface.co/artur-muratov/franx-ner}.} See Appendix~\ref{appendix:seq_lab_details} for all hyperparameter values, training schedule, and post-processing details. 

% \begin{table}[t]
% \centering
% % \small
% \begin{tabular}{lc}
% \hline
% Variant & Exact Match \\
% \hline
% Gold & 34.1 \\
% Unknown & 44.0 \\
% Propagated & \textbf{48.4} \\
% \hline
% \end{tabular}
% \caption{\textbf{Exact match scores (with fuzzy logic) for the BERT-base model on the Gold, Propagated, and Unknown dataset variants for English only, reported in percent.}}
% \label{tab:bert_variant_results}
% \end{table}

\subsubsection{Evaluation}\label{sec:seq_lab_eval}

We evaluated our sequence labeling models using exact span matching complemented by a fuzzy matching mechanism to account for annotation inconsistencies and boundary variations. This framework combines strict character-level alignment with flexible heuristics that capture semantically equivalent predictions, ensuring fair evaluation despite the inherent subjectivity in entity boundary annotation (see Appendix~\ref{app:fuzzy_matching}).

Table~\ref{tab:deberta_comparison} presents exact match scores across the five languages using three DeBERTa configurations. The multilingual DeBERTa-v3-Large model demonstrates consistent performance across all target languages, achieving the best results on three out of five languages (RU: 46.5\%, PT: 66.4\%, BG: 51.6\%). Notably, the English-only model achieves the highest score on English (79.1\%) but shows severe degradation on other languages, with particularly poor performance on Hindi (7.9\%) and Portuguese (28.4\%), confirming the importance of multilingual pretraining for cross-lingual transfer.

 A more detailed evaluation report, including per-language breakdowns and additional metrics, can be found in Appendix~\ref{appendix:evaluation}.

\begin{table}[t]
\centering
\small
\begin{tabularx}{\linewidth}{lXXXXX}
\toprule
Model & EN & RU & HI & PT & BG \\
\midrule
DeBERTa-v3-large-mult & 59.3 & \textbf{46.5} & 40.0 & \textbf{66.4} & \textbf{51.6} \\
mDeBERTa-v3-base-mult & 53.8 & 45.3 & \textbf{52.1} & 62.1 & \textbf{51.6} \\
DeBERTa-v3-large-mono-en & \textbf{79.1} & 45.3 & 7.9 & 28.4 & 35.5 \\
\bottomrule
\end{tabularx}
\caption{\textbf{Sequence labeling performance comparison for coarse-grained role prediction.} Exact match scores (with fuzzy matching) for multilingual and monolingual DeBERTa variants across five languages.}
\label{tab:deberta_comparison}
\end{table}

\subsection{Text Classification}

To obtain fine-grained roles for each span extracted by the sequence labeling model, we use a separate mDeBERTa-v3-Base encoder with a classification head. Each detected span is encoded independently and classified into one or more of 22 detailed roles. The classifier is trained on span-level gold annotations aligned with the same dataset used for entity and coarse-grained role prediction.

\subsubsection{Data Preparation}

To enable multilingual classification, we merge data across five languages and align entity spans using their annotated start and end offsets. For each span, we extract surrounding context to provide the model with localized semantic cues. Specifically, we experiment with various character window sizes and identify optimal classification performance at a symmetric window of 150 characters before the start offset and after the end offset. (see \mbox{Appendix \ref{appendix:textclassification}}). This window provides sufficient contextual information for discriminative role prediction while minimizing noise from unrelated text.

\subsubsection{Model Development}

We formulate fine-grained role prediction as a multi-label classification task conditioned on a high-level role taxonomy. Each training instance includes the entity mention, its surrounding context, and its coarse role. Fine-grained labels are selected from a predefined hierarchy, where each main role (e.g., Protagonist) maps to a subset of allowable fine-grained roles.

The model\footnote{The trained classification model can be found on Hugging Face at  
\url{https://huggingface.co/artur-muratov/franx-cls}.} is trained using binary cross-entropy with logits, selectively masked according to taxonomy constraints:

\[
\mathcal{L} = \frac{\sum_{i} \text{BCE}(y_i, \hat{y}_i) \cdot m_i}{\sum_{i} m_i + \epsilon}
\]

To address class imbalance, we apply class-specific weights during loss computation. These weights are computed from the training distribution using logarithmic scaling of the inverse label frequency.
We demonstrate such full implementation details in Appendix \ref{appendix:textclassification}.

\subsubsection{Evaluation}

We evaluate model predictions using standard classification metrics along with task-specific diagnostics to assess both prediction quality and coverage of fine-grained roles. During inference, the model outputs probability scores over the fine-grained labels and retains roles with confidence greater than 0.01. Finally, we retain scores within 0.05 of the highest-scoring label. This strategy yields a more representative reflection of model uncertainty and score distribution. The thresholds were selected to balance the exact match and recall accuracy.

% We report overall performance in Table \ref{tab:overall_eval_vertical} and per-class performance in Table \ref{tab:lang_eval}. 
% using Micro and Macro F1, Exact Match Accuracy (EMC), and Recall-based Accuracy (REC) 

% \begin{table}[t]
% \centering
% \small
% \begin{tabular}{lc}
% \toprule
% Metric & Score \\
% \midrule
% Micro F1 & 53.0 \\
% Macro F1 & 32.0 \\
% Exact Match Accuracy & 48.0 \\
% Recall Accuracy & 53.0 \\
% \bottomrule
% \end{tabular}
% \caption{Report of overall fine-grained classification performance.}
% \label{tab:overall_eval_vertical}
% \end{table}

% using Exact Match and Recall-based Accuracy:

% \begin{table}[t]
% \centering
% \small
% \begin{tabular}{lcc}
% \toprule
% Language & Exact Match & Recall Accuracy \\
% \midrule
% Portuguese (pt) & 64.7 & 71.6 \\
% Hindi (hi) & 52.9 & 55.7 \\
% Russian (ru) & 47.7 & 53.5 \\
% Bulgarian (bg) & 32.3 & 38.7 \\
% English (en) & 17.6 & 27.5 \\
% \bottomrule
% \end{tabular}
% \caption{Per-language evaluation of fine-grained role predictions.}
% \label{tab:lang_eval}
% \end{table}

%\begin{table}[t]
%\centering
%\scriptsize
%\resizebox{\columnwidth}{!}{%
%\begin{tabular}{lcccc}
%\toprule
%Language & Micro F1 & Macro F1 & Exact Match & Recall Accuracy \\
%\midrule
%All & 53.0 & 32.0 & 48.0 & 53.0 \\
%PT  & —    & —    & \textbf{64.7} & \textbf{71.6} \\
%HI  & —    & —    & 52.9 & 55.7 \\
%RU  & —    & —    & 47.7 & 53.5 \\
%BG  & —    & —    & 32.3 & 38.7 \\
%EN  & —    & —    & 17.6 & 27.5 \\
%\bottomrule
%\end{tabular}%
%}
%\caption{Overall and per-language fine-grained classification results, reported in percent.}
%\label{tab:merged_overall_lang_eval}
%\end{table}

\begin{table}[t]
\small
\centering
\begin{tabular}{ccccccc}
\toprule
\textbf{Lang} & \textbf{Prec} & \textbf{Rec} & \textbf{Mic} & \textbf{Mac}& \textbf{ Acc}& \textbf{EMC}\\
\midrule
All & 51.1& 56.2& 53.5& 31.8& 53.9& 47.8\\
BG  & 34.2& 41.2& 37.3& 23.3& 41.9& 29.1\\
EN  & 23.3& 32.0& 27.0& 19.7& 28.6& 16.5\\
HI  & 58.7& 59.1& 58.9& 28.9& 56.1& 52.5\\
PT  & 69.8& 72.6& 71.2& 24.1& 71.5& 66.4\\
RU  & 48.5& 56.2& 52.1& 22.6& 54.7& 47.6\\
\bottomrule
\end{tabular}
\caption{\textbf{Text classification model results.} We report Precision (Prec), Recall (Rec), Micro F1 (Mic), Marco F1 (Mac), Recall Accuracy (Acc), and Exact Match Accuracy (EMC).}
\label{tab:merged_overall_lang_eval}
\end{table}

 Table~\ref{tab:merged_overall_lang_eval} presents overall and per-language results. The model performs best on frequent and well-covered roles, with an overall micro F1 of 53.5\% and macro F1 of 31.8\%. Portuguese (PT) leads across all metrics, likely due to more consistent annotations and article-level framing, Hindi (HI) and Russian (RU) also show strong performance, while Bulgarian (BG) and English (EN) trail behind. 
 % This could be due to morphological complexity and lexical ambiguity. 

\subsection{Overall System Evaluation}

We evaluate the full system pipeline, which combines entity detection and main role classification (Model 1) with fine-grained role classification (Model 2). Given that our task requires both accurate span identification and contextual role assignment, we report evaluation metrics only over entities that overlap between the predicted spans and the development set. All reported scores in Table~\ref{tab:combined_results} are computed over this overlap set. 

\begin{table}[t]
\small
\centering
\begin{tabular}{ccccccc}
\toprule
\textbf{Lang} & \textbf{Prec} & \textbf{Rec} & \textbf{Mic} & \textbf{Mac}& \textbf{ Acc}& \textbf{EMC}\\
\midrule
All & 48.8 & 54.2 & 51.4 & 31.2& 52.1 & 44.3 \\
BG  & 30.0 & 40.0 & 34.3 & 0.19 & 42.8 & 35.7 \\
EN  & 22.6 & 28.8 & 25.3 & 0.15 & 24.2 & 13.8 \\
HI  & 54.7 & 55.7 & 55.3 & 23.6 & 52.8 & 47.8 \\
PT  & 65.9& 72.5& 69.0& 24.9 & 72.0& 64.0\\
RU  & 50.9 & 58.3 & 54.3 & 25.4 & 55.5 & 42.2 \\
\bottomrule
\end{tabular}
\caption{\textbf{Combined model results.} We report Precision (Prec), Recall (Rec), Micro F1 (Mic), Marco F1 (Mac), Recall Accuracy (Acc), and Exact Match Accuracy (EMC).}
\label{tab:combined_results}
\end{table}

To contextualize our results, we compare our system against three baselines described in \mbox{Appendix \ref{appendix:baselines}}. The results, presented in Table \ref{tab:baseline_comparison}, show that our approach consistently outperforms all baselines by a large margin across all languages. Overall, our Macro F1 of 31.2\,\% surpasses the best baseline (3.4\,\% for B3) by nearly an order of magnitude. Gains are similarly consistent across individual languages.

\begin{table}[t]
\centering
\small
\begin{tabular}{lcccc}\toprule

\textbf{Lang} & \textbf{Ours} & \textbf{Random} & \textbf{Top-k} & \textbf{Freq-Weighted} \\\midrule

EN& \textbf{15.4}& 0.6& 0.6& 3.3\\
HI& \textbf{23.6}& 4.2& 0.1& 1.9\\
RU& \textbf{25.4}& 0.9& 0.0& 2.2\\
BG& \textbf{18.9}& 0.0& 0.0& 4.5\\
PT& \textbf{25.0}& 1.8& 0.0& 3.8\\

 All& \textbf{31.2}& 3.3& 0.2 & 3.4\\ 
 \bottomrule
\end{tabular}
\caption{\textbf{Comparison of system performance against baselines across five languages (Macro F1).}
We compare our systems to three baseline systems, including Random, Top-$k$, and Freq-Weighted.
% B1 indicates Random, B2 indicates Top-$k$, and B3 indicates Freq-Weighted (B3).
}
\label{tab:baseline_comparison}
\end{table}

\subsection{Human Evaluation}

To complement automatic metrics, we conduct a small-scale human evaluation using a custom-built interface\footnote{\url{https://humanevalfranx.streamlit.app/}} (Appendix \ref{appendix:humanannotationplatform}). 
We conducted a human evaluation of role predictions in English, Hindi, and Russian. Annotators were asked to assess the plausibility and confidence of system outputs using a 5-point Likert scale. The annotator pool included both NLP experts and non-expert participants, reflecting a diverse range of use cases.

Table \ref{tab:human_eval} presents the evaluation results. Inter-annotator agreement was measured as the proportion of annotators who selected the majority label for each instance. The overall agreement across all instances was 89.2\%.
English annotations exhibited the highest average confidence, likely reflecting greater fluency and familiarity among annotators. In contrast, the slightly lower confidence and agreement observed for Russian may be attributed to increased ambiguity or a smaller pool of annotators.

\begin{table}[t]
\centering
\small
\begin{tabular}{lcccl}
\toprule
\textbf{Lang} & \textbf{\#Res} & \textbf{\#Anno} & \textbf{AC}  & \textbf{Agr} \\
\midrule
EN & 247& 9& 4.17&88.7\\
HI & 254& 6& 3.75 &93.0\\
RU & 86& 3& 3.60 &80.3\\
\bottomrule
\end{tabular}
\caption{\textbf{Summary of human evaluation across languages.}
Res indicates Responses, Anno indicates Annotators,
AC indicates Average Confidence, and Agr indicates Agreement (\%).
}
\label{tab:human_eval}
\end{table}

\section{Interface}

FRaN-X is an end-to-end framework with multiple functionalities that enable end-users to study entity framing from different perspectives. It enables users to explore framing decisions at both surface and semantic levels. The system is developed using \texttt{Streamlit}, an open-source framework that supports rapid prototyping and seamless deployment of machine learning interfaces. 

Our system is designed with user accessibility and interpretability in mind. The primary users are media researchers, investigative journalists, fact-checkers, and engaged readers who need rapid, cross-lingual framing analysis. It features a modular interface that allows users to explore how entities are portrayed in articles at both a granular and aggregate level.
The system comprises six main pages, each offering distinct functionalities: \textit{Home} (Sec \ref{homepage}), \textit{Analysis} (Sec \ref{analysispage}), \textit{Dynamic Analysis} (Sec \ref{dynamicanalysispage}), \textit{Aggregate Analysis} (Sec \ref{aggregateanalysispage}), \textit{Search} (Sec \ref{searchpage}) and \textit{Timeline} (Sec \ref{timelinepage}).
In addition, an \textit{About} page provides a summary of the system’s features and presents the full taxonomy used for entity role classification.

\subsection{Home Page}
\label{homepage}

The \textit{Home} page enables users to input an article either by pasting its text or by providing a URL. After specifying a filename, users can trigger the entity and role prediction pipeline. The system then displays the detected entities alongside their predicted fine-grained roles in a structured list format (see Appendix \ref{appendix:interface} for an illustration).

\subsection{Analysis Page}
\label{analysispage}

Once an article is processed, it is stored in a session-specific directory (e.g., \verb|session_0|) under the given filename, appended with a timestamp to facilitate tracking multiple uploads.

\begin{figure}
    \centering
    \includegraphics[width=1\linewidth]{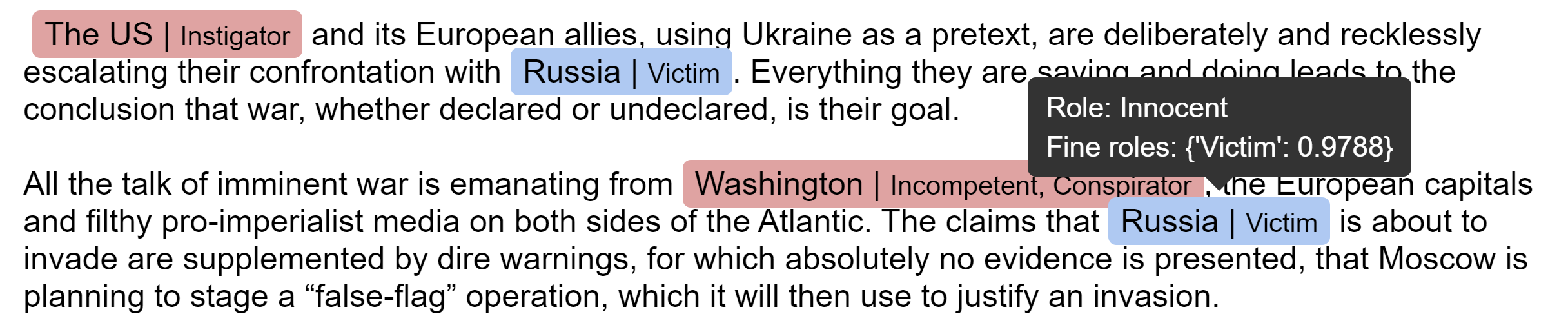}
    \caption{\textbf{An example of the annotated article view.} Additional metadata is displayed on mouse-over.}
    \label{fig:annotated-sentence-with-tooltip}
\end{figure}

Upon selecting an article, users are presented with an annotated view where entities are color-coded according to their coarse-grained roles, and the corresponding fine-grained roles annotated alongside (See Figure~\ref{fig:annotated-sentence-with-tooltip}). Hovering over an entity reveals additional metadata, including the predicted main role and confidence scores for each fine-grained role.

To aid in interpretability, users can toggle the display of repeated annotations. For instance, identical fine-grained roles associated with recurring mentions of the same entity can be hidden or shown depending on user preference. Users can also adjust a confidence threshold slider to filter out low-confidence predictions, thereby controlling the granularity of the information presented.

Further insights are offered through a set of visualizations showing how entities are framed throughout the article (see Appendix \ref{appendix:interface} for an illustration). An additional functionality allows users to extract and view all sentences associated with a specific label, enabling intra-document comparisons of how different entities—or the same entity—are characterized across contexts.

%(see Figures~\ref{fig:dynamic-stacked-bar-chart},~\ref{fig:dynamic-cumulative-bar-chart},~\ref{fig:dynmic-by-main-pie-chart}, and~\ref{fig:dynamic-cumulative-pie-chart})
 
\subsection{Dynamic Analysis Page}
\label{dynamicanalysispage}

This module allows users to load and compare up to four articles side by side. It is particularly useful for examining how different media outlets frame the same event or entity. Examining multiple annotated outputs from different articles may allow users to better understand framing variations across sources (see Appendix \ref{appendix:interface} for an illustration).

\subsection{Aggregate Analysis Page}
\label{aggregateanalysispage}

\begin{figure}[t]
    \centering
    \includegraphics[width=1\linewidth]{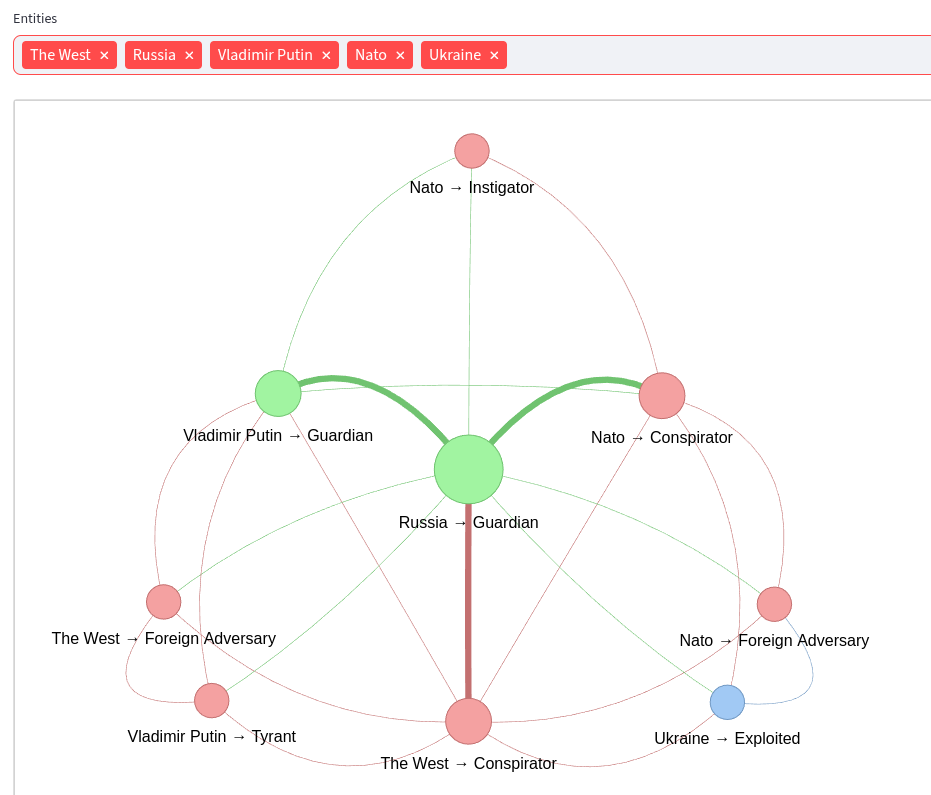}
    \caption{\textbf{Network Graph exploring three articles and the connection between entity, fine-grain role nodes.} Provides the user with an understanding of the cumulative narrative pushed by multiple articles.}
    \label{fig:network-graph}
\end{figure}

Aggregate-level insights are provided both as a standalone page and as part of the dynamic analysis interface. This section includes a network graph that connects entities and their associated roles across multiple documents. Edges between nodes signify co-occurrence within the same article. Users can selectively display nodes and edges based on entity or document preferences. An optional physics-based layout engine enables interactive manipulation of the graph for enhanced usability and interpretation. This is illustrated in Figure~\ref{fig:network-graph}.

\subsection{Search Page}
\label{searchpage}

The \textit{Search} page allows users to select one or more processed articles and search for specific words or phrases (see Appendix \ref{appendix:interface} for an illustration). All matches are highlighted within the full article context. This tool facilitates quick localization of specific entities or terms of interest.

\subsection{Timeline Page}
\label{timelinepage}

The \textit{Timeline} page visualizes the evolution of entity roles across time. For each selected entity, the system displays the sequence of fine-grained roles it has been assigned, along with the confidence scores and surrounding context sentences. In the case of multiple entities, the system also highlights transitions in role assignments (see Appendix \ref{appendix:interface}).

%\begin{figure}
%    \centering
%    \includegraphics[width=1\linewidth]{annotated_sentence.png}
%    \caption{An example of the annotated article view.}
%    \label{fig:annotated-sentence}
%\end{figure}

%\begin{figure}[t]
%    \centering
%    \includegraphics[width=1\linewidth]{dynamic_main_role_bar_chart.png}
%    \caption{\textbf{Stacked Bar Chart comparing main role distributions between articles.}}
%    \label{fig:dynamic-stacked-bar-chart}
%\end{figure}

%\begin{figure}[t]
%    \centering
%    \includegraphics[width=1\linewidth]
%{dynamic_cumulative_fine_grain_bar_chart.png}
%    \caption{\textbf{Cumulative bar chart looking at the main role %distribution broken down into fine-grain roles.}}
%    \label{fig:dynamic-cumulative-bar-chart}
%\end{figure}

%\begin{figure}[t]
%    \centering
%    \includegraphics[width=1\linewidth]{dynamic_cumulative_fine_grain_pie_chart.png}
%   \caption{\textbf{Cumulative pie chart looking at fine-grain role distributions grouped by main role.}}
%    \label{fig:dynmic-by-main-pie-chart}
%\end{figure}

\section{Conclusion and Future Work}
FRaN-X provides end-to-end narrative role detection for multilingual news, addressing the critical limitations of existing media analysis tools that often stop at surface-level sentiment or basic entity recognition. By combining sequence labeling with a unique, fine-grained taxonomy of 22 narrative roles nested under three main categories, FRaN-X empowers researchers, journalists, and the general public to uncover hidden framings and compare how entities are portrayed across sources and languages. Its multilingual, interactive interface makes complex framing insights accessible and actionable.

Looking ahead, we aim to expand FRaN-X’s language coverage to better support low-resource languages, broaden its domain focus to include new topics such as health, politics, and economic crises, and provide an open API to help researchers and policymakers integrate narrative role detection into their workflows. Continued improvements in explainability will further enhance user trust and understanding of how framing decisions are made.

\section{Limitations}

Despite its contributions, FRaN-X relies on manually annotated data, which varies in quality and consistency and can limit the system’s fine-grained role classification performance, especially for languages and domains with sparse or uneven annotations. Some narrative roles remain inherently subjective and ambiguous, which challenges both human annotators and model predictions. Additionally, the system has so far been tested on only five languages, which may limit its generalizability to other linguistic and cultural contexts without further adaptation and evaluation.

\section{Ethics and Broader Impact}

FRaN-X is designed to support transparent, accountable media analysis by revealing how entities are framed in multilingual news narratives. However, its outputs must be interpreted carefully, as misclassification or overreliance on automated role assignments could reinforce biases or misinform public understanding, especially when dealing with sensitive political or social topics. The system’s dependence on training data means it may reflect the biases or gaps present in those sources. We encourage responsible use by journalists, researchers, and policymakers. By releasing the trained models and providing a publicly accessible system, we aim to foster ethical oversight that can advance multilingual narrative analysis while minimizing unintended harms.

\bibliography{custom}

\appendix
\newpage
\clearpage
\section{Data Augmentation Variants}\label{app:data_aug}

\subsection*{Gold}
The \textbf{Gold} variant is the original SemEval 2025
Entity-Framing dataset with no modifications.

\subsection*{Propagated}
The \textbf{Propagated} variant extends gold labels
backwards and forwards within each document.
We apply surface-form matching and alias clustering so that
coreferent mentions (e.g., ``Volodymyr Zelensky'' →
``Zelensky'') inherit the same narrative-role label.
This follows the annotation guideline~\cite{mahmoud2025entityframingroleportrayal},
which states that annotators only need to mark the first clear
mention of an entity; subsequent mentions are optional.

\subsection*{Unknown}
The \textbf{Unknown} variant builds on \textit{Propagated} by additionally
running a spaCy NER pipeline~\cite{honnibal2020spacy} to extract
any named entities not already labeled. These extra spans are given
a dedicated \textit{``Unknown''} class, so the model can learn to distinguish
central, narrative-defining entities from peripheral or background mentions.

\subsection*{Results}

Table~\ref{tab:bert_variant_results} reports the exact-match accuracy 
for the BERT-base baseline on each variant for English only.
The results show that mention propagation improves exact-match accuracy 
by approximately 14.3~percentage points over the Gold variant, while the 
Unknown class does not provide further benefit and slightly reduces final performance.

\begin{table}[h]
\centering
\begin{tabular}{lc}
\hline
Variant & Exact Match \\
\hline
Gold & 34.1 \\
Unknown & 44.0 \\
Propagated & \textbf{48.4} \\
\hline
\end{tabular}
\caption{\textbf{Exact match scores (with fuzzy logic) for the BERT-base baseline.}}
\label{tab:bert_variant_results}
\end{table}

\section{Sequence Labeling Details}
\label{appendix:seq_lab_details}

The final backbone is DeBERTa-v3-Large, which has 24 Transformer layers, a hidden size of 1024, 16 heads, and disentangled relative attention. We retain the original SentencePiece vocabulary and split long articles into overlapping windows of 1024 tokens with a stride of 256. Each contextualized token vector is fed into a linear classification head to produce seven logits for \{O, B/I-Protagonist, B/I-Antagonist, B/I-Innocent\}. To address class imbalance, we add a bias of +0.2 to each non-O logit at initialization and apply a +1.0 shift at inference.

We use DeBERTa’s default dropout of 0.1 on hidden states and attention probabilities. The training loss is the negative log-likelihood of the gold tag sequence, masking padding tokens. Training runs for 20 epochs with AdamW and a cosine learning rate schedule (initial learning rate \(1 \times 10^{-5}\) with 10\% warm-up). The batch size is 2 due to the long sequences, and non-O labels are upweighted by a factor of 2.0 in the loss. The best checkpoint is selected by dev-set F1 score.

The CRF output is post-processed using the tokenizer’s \texttt{offset\_mapping} to reconstruct character-level spans. A custom \texttt{ScaledSpanMerger} (threshold 0.5) merges fragmented tokens, while spans consisting only of punctuation, stop-words, or fewer than two characters are discarded.

\section{Fuzzy Matching}
\label{app:fuzzy_matching}

The primary evaluation metric is exact match accuracy, calculated as the percentage of gold entity spans that have a corresponding predicted span with the same character boundaries and coarse role label. To complement this strict metric, we compute standard NER evaluation measures including precision, recall, and F1-score at both the span level and the token level using the \texttt{seqeval}~\cite{seqeval} framework.

Given the inherent subjectivity in entity boundary annotation, especially for complex named entities and descriptive phrases, we implement a fuzzy matching mechanism that considers a prediction correct if it satisfies any of the following conditions: (1) Exact match: identical character spans and role labels; (2) String similarity: after normalizing punctuation and whitespace, the lowercased spans are identical; (3) Acronym matching: one span is a valid acronym of the other (e.g., "UN" matches "United Nations"); (4) Substring containment: one span contains the other as a substring, with a minimum length threshold of 3 characters; (5) Token overlap: substantial word-level overlap ($\geq 67\%$ of tokens) between spans of the same role; (6) Character overlap: at least 80\% character-level overlap between the predicted and gold spans.

The fuzzy matching mechanism proves essential for fair evaluation, as strict character-level matching would penalize semantically correct predictions with minor boundary differences. Our analysis shows that approximately 15-25\% of the matches across languages are captured only through fuzzy logic, highlighting the importance of this evaluation approach for real-world entity recognition tasks.

\section{Detailed Evaluation Reports}\label{appendix:evaluation}

As presented in Table~\ref{tab:combined_eval}, we report three main types of metrics for our sequence labeling models: overall accuracy, per-role precision, recall, and F1, and macro/micro averages. Accuracy is computed as the proportion of gold entity spans that are correctly matched by a predicted span with the same boundaries and coarse role label, using fuzzy matching to allow for minor boundary differences. For precision and recall, we follow the annotation guideline that only the first mention of each entity with a given role is required to be annotated, and repeated mentions are optional. To ensure fair evaluation, we apply a deduplication step: for each unique entity surface form (normalized for whitespace and case) and role, we count at most one true positive and one false positive, regardless of how many times the model predicts that entity in the document. This prevents the model from being penalized for predicting multiple mentions of the same entity when only the first is annotated. Recall is calculated as the proportion of unique gold entity-role pairs that are matched by at least one prediction, while precision is the proportion of unique predicted entity-role pairs that match a gold annotation. False negatives are counted for each unique gold entity-role pair not matched by any prediction, and false positives for each unique predicted entity-role pair not present in the gold annotations. Macro and micro averages are computed over the three main narrative roles (Protagonist, Antagonist, Innocent). This evaluation protocol ensures that our metrics reflect both the annotation policy and the practical requirements of entity-centric narrative role recognition.

% Use table* if you need a wide table in a two-column paper:
\begin{table*}[t]
\centering
\scriptsize
\setlength{\tabcolsep}{4pt} % tighten column padding
\resizebox{\textwidth}{!}{%
  \begin{tabular}{ll
    ccc   ccc   ccc   ccc}
  \toprule
  \textbf{Lang.} & \textbf{Model} 
    & \multicolumn{3}{c}{\textbf{Antagonist}} 
    & \multicolumn{3}{c}{\textbf{Innocent}} 
    & \multicolumn{3}{c}{\textbf{Protagonist}} 
    & \multicolumn{3}{c}{\textbf{Totals / Accuracy}} \\
  \cmidrule(lr){3-5}\cmidrule(lr){6-8}\cmidrule(lr){9-11}\cmidrule(lr){12-14}
   & 
    & P    & R    & F1   
    & P    & R    & F1   
    & P    & R    & F1   
    & \#/\%            & W.\,Acc & U.\,Acc \\
  \midrule
  RU & DeBERTa-v3-large-mult   
    & 33.3 & 47.9 & 39.3 
    & 10.0 & 11.1 & 10.5 
    & 29.4 & 53.6 & 38.0 
    & 40/86 (46.5\%) & 46.5    & 48.5    \\
     & mDeBERTa-v3-base-mult    
    & 31.7 & 54.2 & 40.0 
    &  8.3 & 11.1 &  9.5 
    & 23.8 & 37.0 & 29.0 
    & 39/86 (45.3\%) & 45.3    & 43.4    \\
     & DeBERTa-v3-large-mono-en 
    & 15.6 & 56.5 & 24.4 
    &  5.3 & 22.2 &  8.5 
    & 12.3 & 28.6 & 17.2 
    & 39/86 (45.3\%) & 45.3    & 38.4    \\
  \midrule
  EN & DeBERTa-v3-large-mult   
    & 32.2 & 60.3 & 42.0 
    & 16.7 & 37.5 & 23.1 
    &  8.0 & 22.2 & 11.8 
    & 54/91 (59.3\%) & 59.3    & 60.5    \\
     & mDeBERTa-v3-base-mult    
    & 27.7 & 57.1 & 37.3 
    & 10.5 & 22.2 & 14.3 
    &  2.6 & 11.1 &  4.3 
    & 49/91 (53.8\%) & 53.9    & 58.8    \\
     & DeBERTa-v3-large-mono-en 
    & 22.0 & 81.0 & 34.6 
    & 13.6 & 75.0 & 23.1 
    &  6.7 & 44.4 & 11.6 
    & 72/91 (79.1\%) & 79.1    & 79.8    \\
  \midrule
  HI & DeBERTa-v3-large-mult   
    & 20.9 & 27.3 & 23.6 
    & 11.7 & 13.8 & 12.7 
    & 19.8 & 55.6 & 29.2 
    & 112/280 (40.0\%) & 40.0    & 41.0    \\
     & mDeBERTa-v3-base-mult    
    & 20.4 & 48.3 & 28.7 
    & 18.3 & 30.6 & 22.9 
    & 22.9 & 56.2 & 32.5 
    & 146/280 (52.1\%) & 52.1    & 54.0    \\
     & DeBERTa-v3-large-mono-en 
    & 13.8 & 16.0 & 14.8 
    &  5.9 &  1.5 &  2.4 
    & 25.0 &  3.4 &  6.0 
    &  22/280 (7.9\%) &  7.9    &  8.8    \\
  \midrule
  PT & DeBERTa-v3-large-mult   
    & 40.6 & 48.1 & 44.1 
    & 43.1 & 60.8 & 50.4 
    & 27.4 & 82.1 & 41.1 
    &  77/116 (66.4\%) & 66.4    & 71.4    \\
     & mDeBERTa-v3-base-mult    
    & 31.0 & 48.1 & 37.7 
    & 42.0 & 58.0 & 48.7 
    & 28.8 & 67.9 & 40.4 
    &  72/116 (62.1\%) & 62.1    & 64.4    \\
     & DeBERTa-v3-large-mono-en 
    & 26.2 & 59.3 & 36.4 
    &  0.0 &  0.0 &  0.0 
    & 20.0 & 53.6 & 29.1 
    &  33/116 (28.4\%) & 28.5    & 33.7    \\
  \midrule
  BG & DeBERTa-v3-large-mult   
    & 32.3 & 55.6 & 40.8 
    & 16.7 & 66.7 & 26.7 
    & 22.2 & 25.0 & 23.5 
    &  16/31 (51.6\%) & 51.6    & 55.0    \\
     & mDeBERTa-v3-base-mult    
    & 19.5 & 42.1 & 26.7 
    &  7.7 & 33.3 & 12.5 
    & 21.1 & 66.7 & 32.0 
    &  16/31 (51.6\%) & 51.6    & 51.7    \\
     & DeBERTa-v3-large-mono-en 
    & 15.9 & 52.6 & 24.4 
    &  0.0 &  0.0 &  0.0 
    &  0.0 &  0.0 &  0.0 
    &  11/31 (35.5\%) & 35.5    & 38.3    \\
  \bottomrule
  \end{tabular}%
}
\caption{Span-level metrics and overall accuracy for all language–model pairs, reported in percent except raw counts.}
\label{tab:combined_eval}
\end{table*}

\section{Text Classification Details}\label{appendix:textclassification}

We fine-tuned a taxonomy-aware multi-label classification model using the HuggingFace Transformers framework. All experiments were conducted on a single NVIDIA Quadro RTX 6000 GPU with 24 GB of memory. The following configurations were used:

\begin{itemize}
    \item \textbf{Model:} mDeberta-v3-base
    \item \textbf{Learning Rate:} 2e-5
    \item \textbf{Batch Size:} 3 
    \item \textbf{Number of Epochs:} 10
    \item \textbf{Weight Decay:} 0.01
    \item \textbf{Evaluation Strategy:} Every epoch
    \item \textbf{Early Stopping:} Enabled, with patience of 3 epochs
    \item \textbf{Random Seed:} 42
\end{itemize}

\subsection{Context window experimentation}

We experiment with various character window sizes of 75, 150 and 180. The results for exact match accuracy are in Table~\ref{tab:context_size}.

\begin{table}
    \centering
    \begin{tabular}{cc}\toprule
         Context Window size& Exact Match Accuracy\\\midrule
         75& 46.6\\
         150& 48.0\\
 180&42.1\\ \bottomrule
    \end{tabular}
    \caption{Exact match accuracy for context window sizes, reported in percent}
    \label{tab:context_size}
\end{table}

\section{Baselines}\label{appendix:baselines}

As a quick overview, the performance of our three random baselines is summarized in 
Table~\ref{tab:random_baseline}, Table~\ref{tab:topk_baseline}, and Table~\ref{tab:weighted_baseline}.

\subsection*{Baseline 1: Fully Random Prediction}

To establish a true chance‐level lower bound, we first implement a “fully random” predictor. For each test instance, we sample the number of fine‐grained role labels k from the empirical distribution of label counts observed in the training data (e.g. ~90\% of examples have one label, ~10\% have two, etc.). We then select k labels uniformly at random from the universe of all possible roles. Because this baseline ignores both marginal label frequencies and co‐occurrence structure, it quantifies the performance one would expect in the absence of any learned signal.

\begin{table}[t]
  \centering
  \scriptsize
  \resizebox{0.5\textwidth}{!}{%
    \begin{tabular}{lcccccc}
      \toprule
      Lang. &    P &    R & Micro F1 & Macro F1 & Recall Acc & Exact Match Acc \\
      \midrule
      bg   &  0.0 &  0.0 &   0.0    &   0.0    &   0.0      &   0.0           \\
      en   &  1.7 &  1.5 &   1.6    &   0.6    &   1.7      &   1.7           \\
      hi   &  7.0 &  6.4 &   6.7    &   4.2    &   5.6      &   5.6           \\
      pt   &  2.7 &  2.5 &   2.6    &   1.8    &   2.7      &   2.7           \\
      ru   &  2.2 &  2.1 &   2.2    &   0.9    &   0.0      &   0.0           \\
      All  &  4.2 &  3.8 &   4.0    &   3.3    &   3.3      &   3.3           \\
      \bottomrule
    \end{tabular}%
  }
  \caption{Random baseline metrics, reported in percent.}
  \label{tab:random_baseline}
\end{table}

\subsection*{Baseline 2: Top-k Frequent Labels}

Next, we test a simple majority‐class strategy. Again sampling k from the training label‐count distribution, this baseline always predicts the k most frequent labels in the training set for that language (e.g. “Victim,” “Foreign Adversary,” etc.). This approach serves as a strong “null” competitor whenever a few labels dominate; poor performance here indicates the test split does not simply reuse the train’s most common roles.

\begin{table}[t]
  \centering
  \resizebox{0.5\textwidth}{!}{%
    \begin{tabular}{lcccccc}
      \toprule
      Lang. &    P &    R & Micro F1 & Macro F1 & Recall Acc & Exact Match Acc \\
      \midrule
      bg   &  0.0 &  0.0 &   0.0    &   0.0    &   0.0      &   0.0           \\
      en   &  6.9 &  6.1 &   6.5    &   0.6    &   5.2      &   5.2           \\
      hi   &  1.4 &  1.3 &   1.3    &   0.1    &   1.4      &   1.4           \\
      pt   &  0.0 &  0.0 &   0.0    &   0.0    &   0.0      &   0.0           \\
      ru   &  0.0 &  0.0 &   0.0    &   0.0    &   0.0      &   0.0           \\
      All  &  1.8 &  1.6 &   1.7    &   0.2    &   1.5      &   1.5           \\
      \bottomrule
    \end{tabular}%
  }
  \caption{Top-\textit{k} baseline metrics, reported in percent.}
  \label{tab:topk_baseline}
\end{table}

\subsection*{Baseline 3: Frequency-Weighted Sampling}

Finally, we introduce a frequency‐weighted random predictor that still respects the global popularity of each role but allows diversity beyond the top few. After sampling k from the training label‐count distribution, we draw k distinct labels without replacement, where each label’s selection probability is proportional to its marginal frequency in the training data. This baseline therefore captures the benefit of knowing which roles are generally common, while remaining agnostic to any label co‐occurrence patterns.

\begin{table}[t]
  \centering
  \resizebox{0.5\textwidth}{!}{%
    \begin{tabular}{lcccccc}
      \toprule
      Lang. &    P &    R & Micro F1 & Macro F1 & Recall Acc & Exact Match Acc \\
      \midrule
      bg   &  7.1  &  6.7  &   6.9    &   4.5    &   0.0      &   0.0           \\
      en   &  3.4  &  3.0  &   3.2    &   3.3    &   1.7      &   1.7           \\
      hi   &  2.8  &  2.6  &   2.7    &   1.9    &   1.4      &   1.4           \\
      pt   &  6.7  &  6.3  &   6.5    &   3.8    &   5.3      &   5.3           \\
      ru   &  4.4  &  4.2  &   4.3    &   2.2    &   4.4      &   4.4           \\
      All  &  4.2  &  3.8  &   4.0    &   3.4    &   2.7      &   2.7           \\
      \bottomrule
    \end{tabular}%
  }
  \caption{Frequency-Weighted Baseline Metrics, reported in percent (one decimal).}
  \label{tab:weighted_baseline}
\end{table}
\section{Entity Framing Taxonomy}
\label{appendix:taxonomy}
The entity framing taxonomy is based on the work of \cite{mahmoud2025entityframingroleportrayal} and we present in Figure~\ref{fig:st1_taxonomy}.

\begin{figure}[t]
\small
\centering
\begin{minipage}[t]{0.3\linewidth}
\textbf{Protagonist}\\
\begin{enumerate}[nosep,leftmargin=*]
    \item Guardian
    \item Martyr
    \item Peacemaker
    \item Rebel
    \item Underdog
    \item Virtuous
\end{enumerate}
\end{minipage}
\hfill
\begin{minipage}[t]{0.3\linewidth}
\textbf{Antagonist}\\
\begin{enumerate}[nosep,leftmargin=*]
    \item Instigator
    \item Conspirator
    \item Tyrant
    \item Foreign Adversary
    \item Traitor
    \item Spy
    \item Saboteur
    \item Corrupt
    \item Incompetent
    \item Terrorist
    \item Deceiver
    \item Bigot
\end{enumerate}
\end{minipage}
\hfill
\begin{minipage}[t]{0.3\linewidth}
\textbf{Innocent}\\
\begin{enumerate}[nosep,leftmargin=*]
    \item Forgotten
    \item Exploited
    \item Victim
    \item Scapegoat
\end{enumerate}
\end{minipage}
\caption{Entity framing taxonomy.}
\label{fig:st1_taxonomy}
\end{figure}

\section{Error Analysis}\label{appendix:error_analysis}

To gain deeper insights into the performance of our models, particularly around fine-grained role predictions, we conducted a focused error analysis. To facilitate rapid and scalable manual inspection, we developed an internal error analysis interface using Streamlit. This diagnostic interface enabled us to efficiently explore example annotations across languages, gold fine-grained roles, and predicted labels. A sample snapshot of this interface is shown in Figure~\ref{fig:erroranalysis-UI}

During these investigations, we identified two consistent issues: 
1. For many entity mentions, especially in English, the model did not output any fine-grained roles. This contributed to a very low exact match accuracy initially. For example, in early experiments, the English exact match accuracy was as low as 5\%. Although threshold tuning helped marginally, a significant number of entities remained unassigned.
2. In instances where gold annotations contained multiple fine-grained roles, the model often predicted only one. To better capture partial correctness in such cases, we computed recall-based accuracy, which considers a prediction to be correct if at least one gold role is retrieved. This metric revealed stronger performance, particularly for English.

In response to these observations, we refined our prediction strategy. Instead of relying solely on a fixed classification threshold, we adopted a margin-based decoding scheme. We retained all roles with predicted confidence above 0.01 and then selected those within a 0.05 margin of the highest scoring role. The margin 0.05 was selected to balance exact match accuracy and recall based accuracy. This ensured that the model always made at least one prediction.

To promote transparency and informed use, we also display predicted confidence scores in the user interface. This decision was grounded in both usability and ethical considerations as users should be aware of the model's uncertainty. This is especially integral in socially sensitive applications where an incorrectly predicted role can reinforce bias or propagate misinformation.

Despite these improvements, manual inspection of predicted annotations revealed that many low confidence predictions were semantically appropriate in the given context. These patterns, along with moderate inter-annotator agreement on gold annotations, underscore the inherently subjective nature of the task.

These observations motivated our small-scale humane valuation that aimed to understand how human annotators perceive the model's predicted fine-grained roles given the appropriate context.

% \begin{figure*}
%     \centering
%     \includegraphics[width=1\linewidth]{error_analysis.png}
%     \caption{\textbf{Error Analysis internal interface}}
%     \label{fig:erroranalysis-UI}
% \end{figure*}

% \newpage
% \clearpage
\section{Human Annotation Platform}\label{appendix:humanannotationplatform}

We conducted a human evaluation to assess the quality of our model’s predicted fine-grained roles across five languages. We use the development set across all five languages and divide each language into manageable segments. To balance annotation effort across languages, we varied the number of segments based on entity volume, annotator availability, and article distribution. 

High-volume languages like Hindi were split into smaller batches for manageability, while languages with fewer entities, such as Portuguese and Bulgarian, used fewer segments due to lack of annotators. Some final counts (e.g., English Segment 5 with only 4 entities) reflect residuals from uneven article-level distribution. Rather than enforcing strict segment uniformity, we prioritized preserving natural document boundaries and annotator pacing. 

We summarize the distribution of annotated entity mentions across all segments and languages used in the human evaluation (Table {tab:entity\_counts\}). This helps quantify the density and segmentation of evaluation instances. 

\begin{table}[t]
\centering
\small
\begin{tabular}{l l r}
\toprule
\textbf{Language} & \textbf{Segment} & \textbf{Num Entities} \\
\midrule
bg & Segment 1 & 14 \\
pt & Segment 1 & 75 \\
hi & Segment 1 & 27 \\
hi & Segment 2 & 28 \\
hi & Segment 3 & 12 \\
hi & Segment 4 & 25 \\
hi & Segment 5 & 26 \\
hi & Segment 6 & 24 \\
ru & Segment 1 & 15 \\
ru & Segment 2 & 14 \\
ru & Segment 3 & 14 \\
ru & Segment 4 & 2 \\
en & Segment 1 & 15 \\
en & Segment 2 & 15 \\
en & Segment 3 & 11 \\
en & Segment 4 & 13 \\
en & Segment 5 & 4 \\
\bottomrule
\end{tabular}
\caption{Entity Count by Segment Across All Languages}
\label{tab:entity_counts}
\end{table}

The annotators assess model predictions across three languages (English, Russian, and Hindi). Each annotator is shown the article, the target entity, its main role, and the model-generated fine-grained roles. For each predicted role, annotators indicate whether the role is correct (appropriate in context), incorrect, or unclear. Additionally, they rate their confidence on a 1–5 scale. 

Annotators include students from undergraduate and graduate programs with backgrounds in natural language processing and linguistics, along with a small number of general public annotators. This diverse selection of annotators ensures a balance between domain expertise and layperson interpretability, allowing us to assess both technical alignment and broader applicability of predicted roles understanding. We observe high inter-annotator agreement across these groups in Appendix \ref{appendix:humanannotationplatform}, indicating that model predictions are intelligible and broadly interpretable. 

\begin{figure*}
    \centering
    \includegraphics[width=1\linewidth]{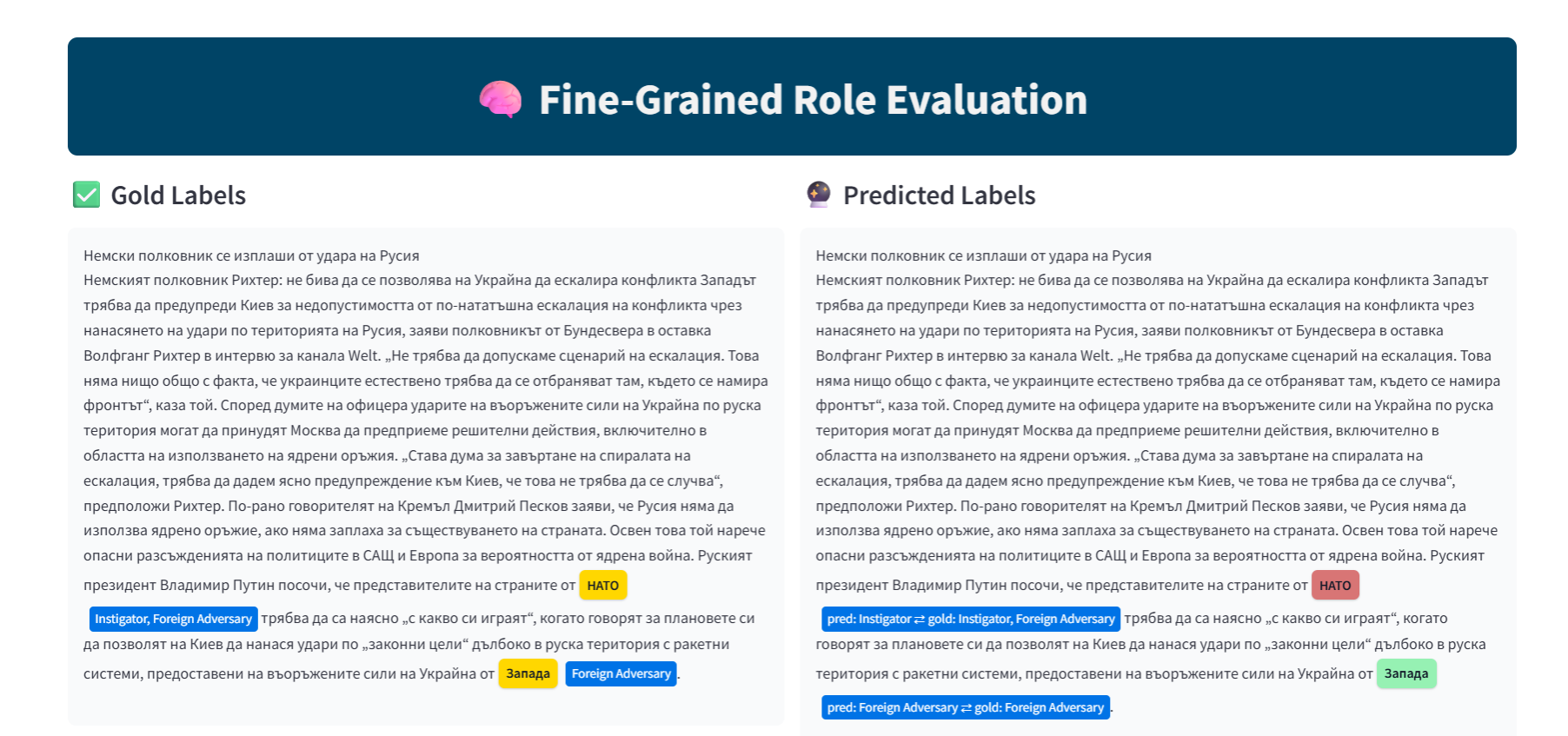}
    \caption{\textbf{Error Analysis internal interface}}
    \label{fig:erroranalysis-UI}
\end{figure*}

\begin{figure*}
    \centering
    \includegraphics[width=1\linewidth]{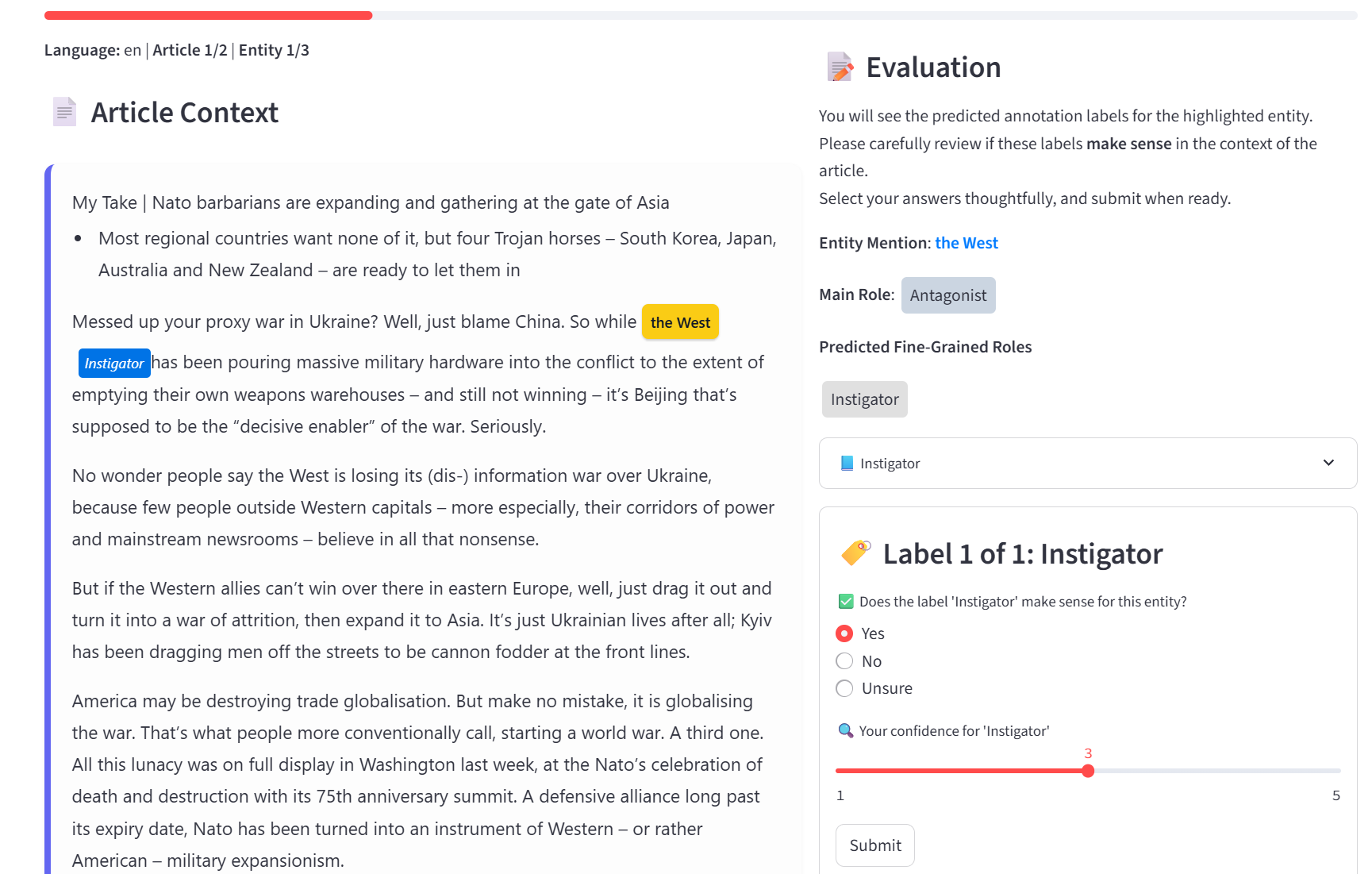}
    \caption{Human evaluation tool deployed on Streamlit: \url{https://humanevalfranx.streamlit.app/}.}
    \label{Human Eval UI}
\end{figure*}

% \begin{table}[h]
% \small
% \centering
% \begin{tabular}{|p{4cm}|p{5cm}|p{4cm}|}
% \hline
% \textbf{1. Protagonist} & \textbf{2. Antagonist} & \textbf{3. Innocent} \\ \hline
% \begin{enumerate} [nosep]
%     \item Guardian
%     \item Martyr
%     \item Peacemaker
%     \item Rebel
%     \item Underdog
%     \item Virtuous
% \end{enumerate} 
% &
% \begin{enumerate} [nosep]
%     \item Instigator
%     \item Conspirator
%     \item Tyrant
%     \item Foreign Adversary
%     \item Traitor
%     \item Spy
%     \item Saboteur
%     \item Corrupt
%     \item Incompetent
%     \item Terrorist
%     \item Deceiver
%     \item Bigot
% \end{enumerate} 
% &
% \begin{enumerate} [nosep]
%     \item Forgotten
%     \item Exploited
%     \item Victim
%     \item Scapegoat
% \end{enumerate} \\ \hline
% \end{tabular}
% \caption{Entity Framing Taxonomy at a Glance.}
% \label{tab:st1_taxonomy}

\newpage
\clearpage
\onecolumn

\section{Interface Visualizations}\label{appendix:interface}
% \begin{figure}[t]
%     \centering
%     \includegraphics[width=1\linewidth]{}
\begin{figure}[th]
        \centering
        \includegraphics[width=0.805\linewidth]{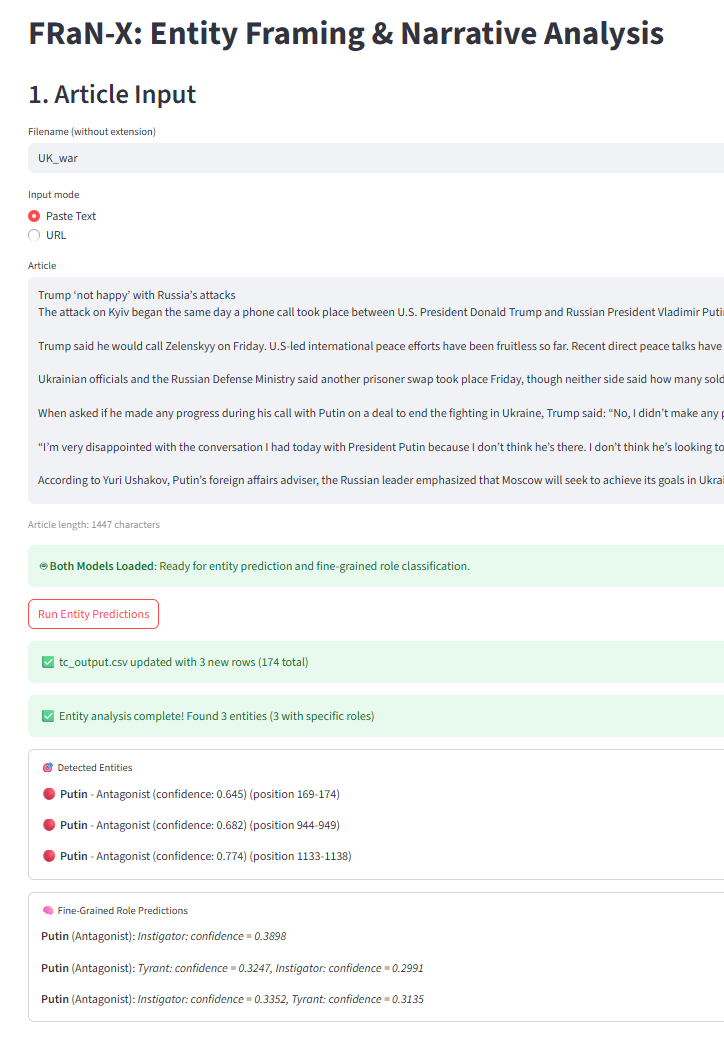}
               \caption{\textbf{Home page of the FRaN-X interface}}
    \label{fig:homepage}
    \end{figure}

\begin{figure*}
    \centering
    \includegraphics[width=1\linewidth]{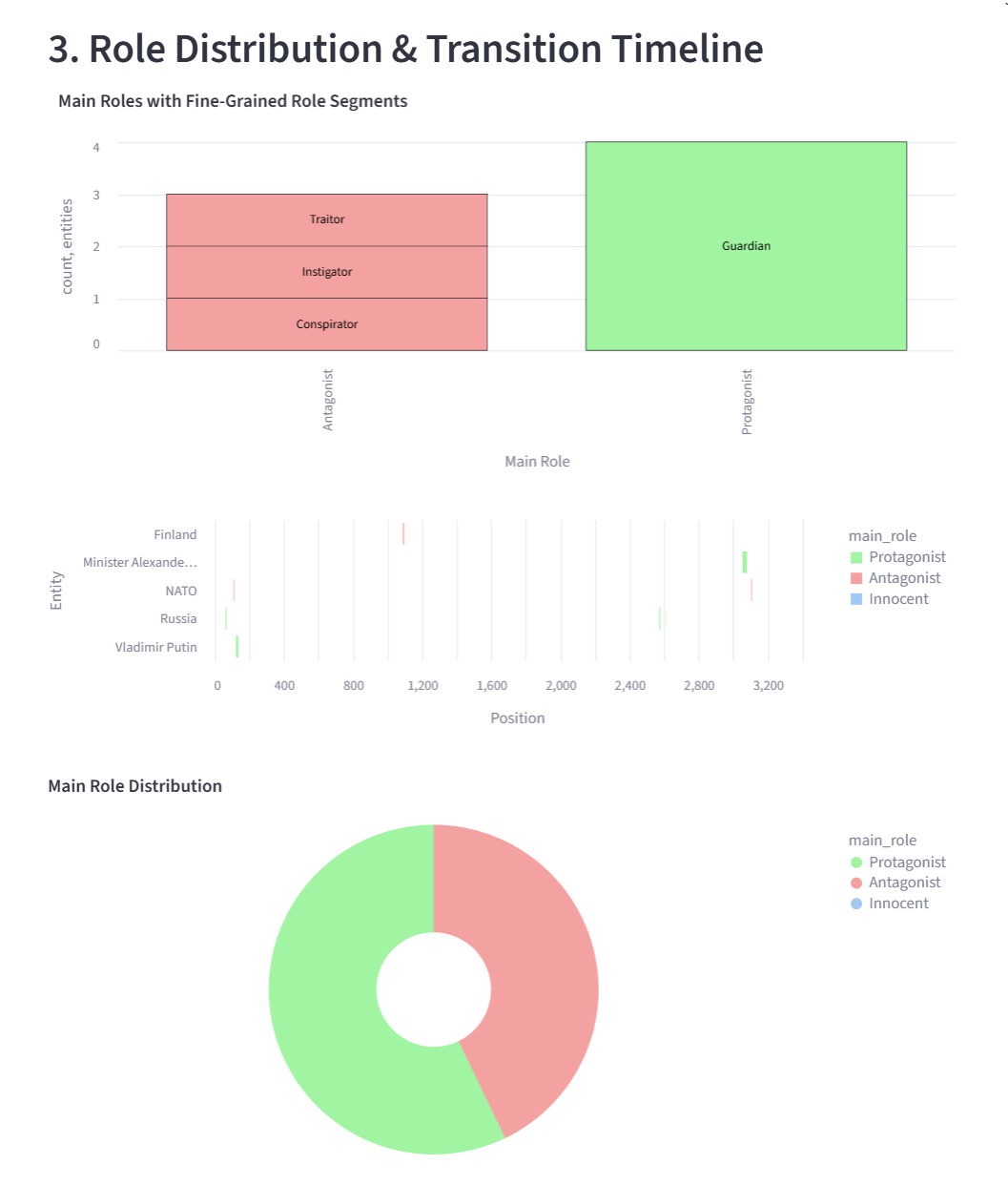}
    \caption{\textbf{Main Role Distribution Data Visualization and Entity Timeline.} Through these charts users can better understand main role distributions and how it is referenced throughout the article.}
    \label{fig:analysis-charts}
\end{figure*}

\begin{figure*}
    \centering
    \includegraphics[width=1\linewidth]{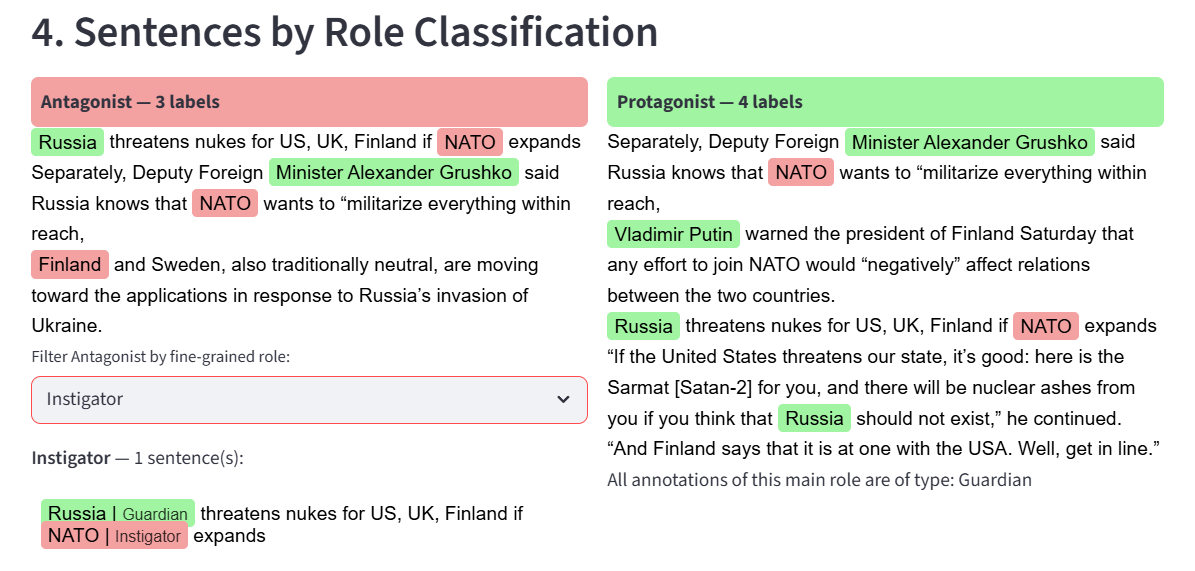}
    \caption{\textbf{Sentence Extraction.} Allows for users to extract all the sentences of a chosen main or fine grain role.}
    \label{fig:enter-label}
\end{figure*}

%\begin{figure*}[t]
%    \centering
%    \includegraphics[width=1\linewidth]{dynamic_fine_grain_role_pie_chart.png}
%    \caption{Cumulative pie chart looking at fine grain role distributions}
%    \label{fig:dynamic-cumulative-pie-chart}
%\end{figure*}

\begin{figure*}
        \centering
        \includegraphics[width=1\linewidth]{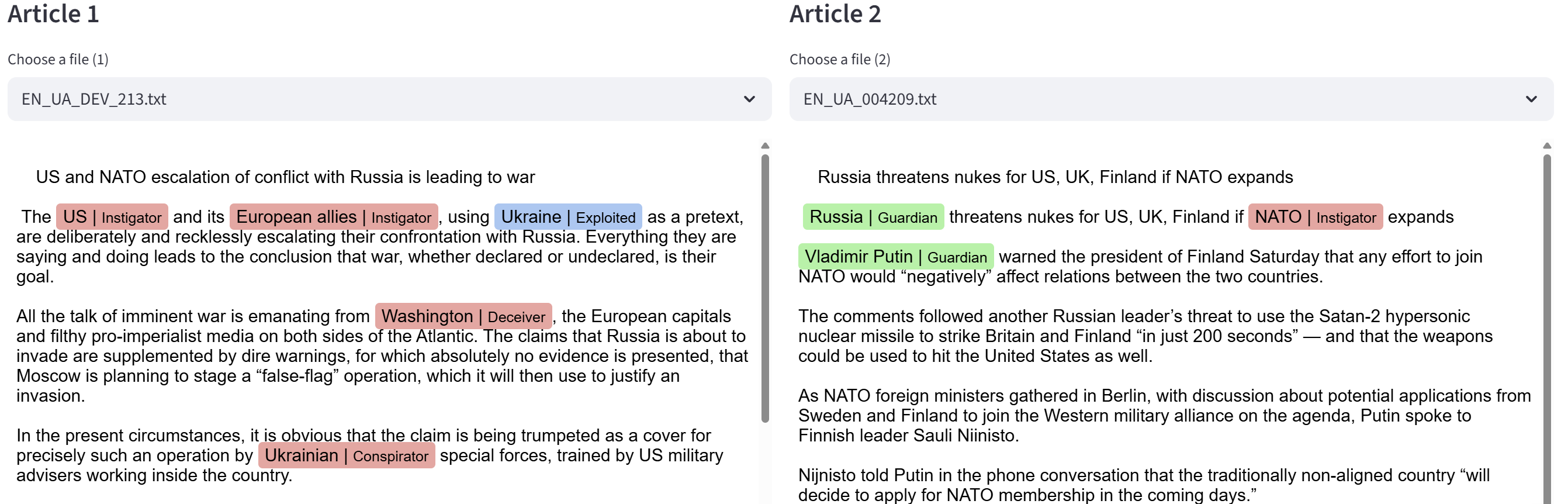}
               \caption{\textbf{Side-by-Side Article Comparison.} FRaN-X supports side-by-side comparisons for up to 4 articles.}
    \label{fig:homepage}
\end{figure*}

\begin{figure*}[h]
    \centering
    \includegraphics[width=1\linewidth]{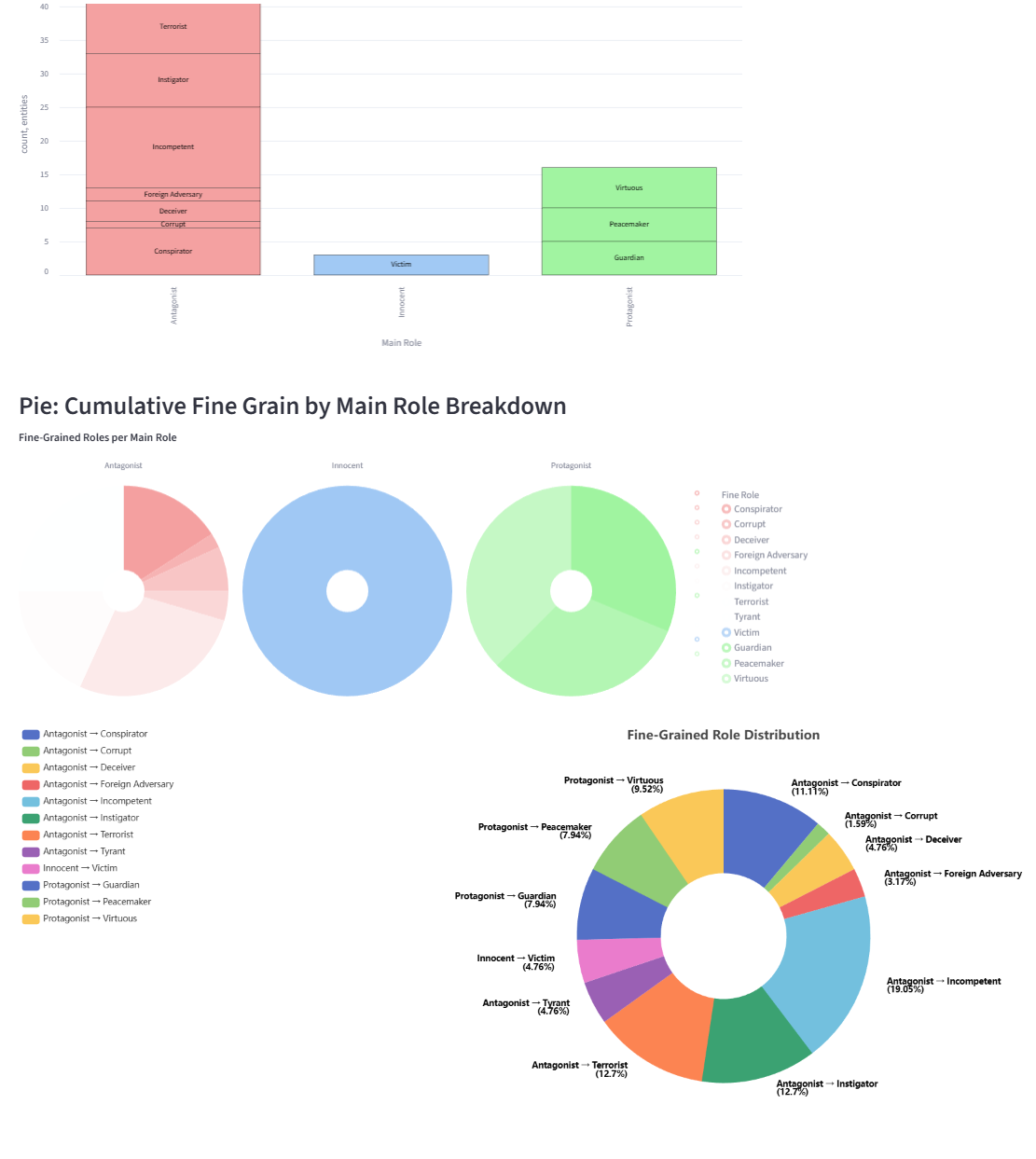}
    \caption{\textbf{Cumulative Entity Distributions.} Users can analyze the cumulative distributions of multiple articles at once in the Dynamic Analysis Page.}
    \label{fig:dynamic-cumulative-distributions}
\end{figure*}

\begin{figure*}
    \centering
    \includegraphics[width=1\linewidth]{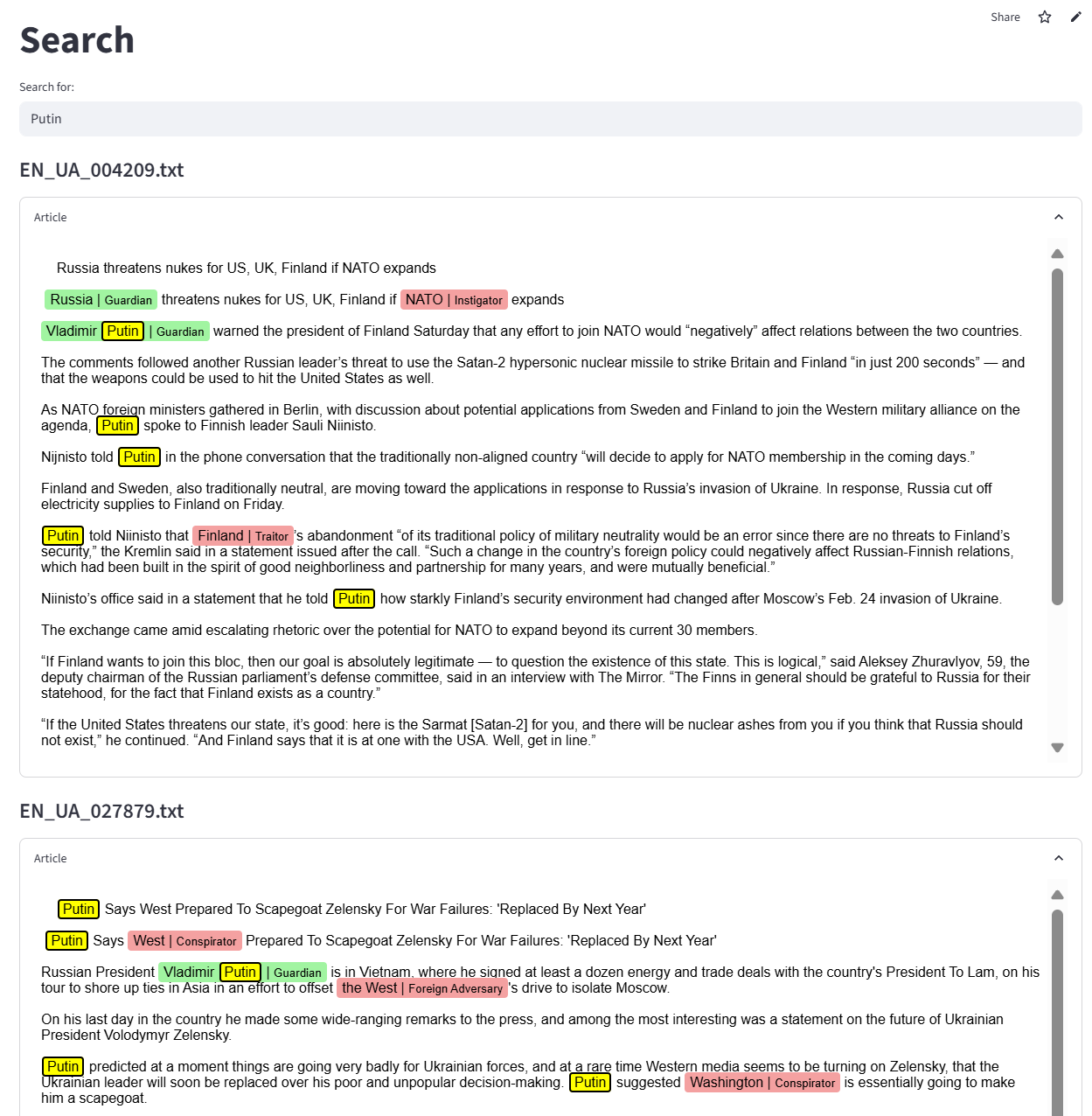}
    \caption{\textbf{Search Within a Set of Articles.} User can search for instances of a specific word.}
    \label{fig:search-page}
\end{figure*}

\begin{figure*}
    \centering
    \includegraphics[width=1\linewidth]{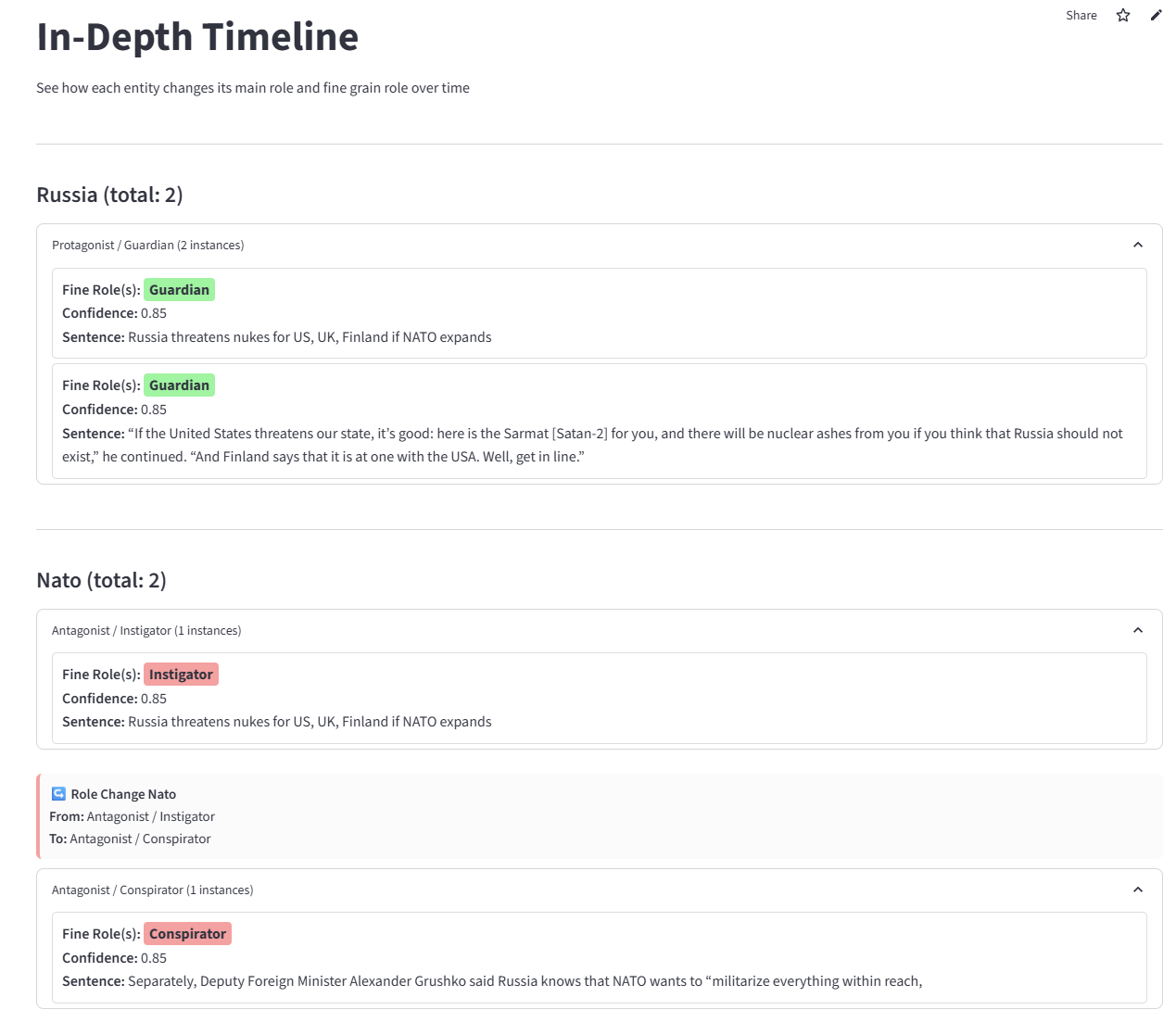}
    \caption{\textbf{Entity Timeline.} Users can track how entities are framed over the course of an article.}
    \label{fig:enter-label}
\end{figure*}

% \end{table}

\end{document}